\documentclass[10pt, a4paper]{article}
\usepackage{lrec2022} 
\usepackage{multibib}
\newcites{languageresource}{Language Resources}
\pdfoutput=1
\usepackage[pdftex]{graphicx}
\usepackage{tabularx}
\usepackage{soul}

\usepackage{titlesec}
\titleformat{\section}{\normalfont\large\bfseries\center}{\thesection.}{1em}{}
\titleformat{\subsection}{\normalfont\SmallTitleFont\bfseries\raggedright}{\thesubsection.}{1em}{}
\titleformat{\subsubsection}{\normalfont\normalsize\bfseries\raggedright}{\thesubsubsection.}{1em}{}
\renewcommand\thesection{\arabic{section}}
\renewcommand\thesubsection{\thesection.\arabic{subsection}}
\renewcommand\thesubsubsection{\thesubsection.\arabic{subsubsection}}

\usepackage{epstopdf}
\usepackage[utf8]{inputenc}

\usepackage{hyperref}
\usepackage{xstring}

\usepackage{color}

\usepackage{tabularx}
\usepackage{url}
\usepackage{booktabs}
\usepackage{multirow}
\usepackage{colortbl}
\usepackage{subcaption}
\usepackage{caption}
\usepackage{CJKutf8}
\usepackage{titling}

\title{Annotation-Scheme Reconstruction for ``Fake News''\\and Japanese Fake News Dataset}

\name{Taichi Murayama\textsuperscript{\rm *}$^{1}$, Shohei Hisada\textsuperscript{\rm *}$^{2}$, Makoto Uehara$^{2}$, Shoko Wakamiya$^{2}$, Eiji Aramaki$^{2}$}

\address{$^1$SANKEN Osaka University,\\
$^2$NARA Institute of Science and Technology, \\
taichi88@sanken.osaka-u.ac.jp, \{s-hisada, uehara.makoto.ug2, wakamiya, aramaki\}@is.naist.jp\\}

\abstract{
Fake news provokes many societal problems; therefore, there has been extensive research on fake news detection tasks to counter it.
Many fake news datasets were constructed as resources to facilitate this task.
Contemporary research focuses almost exclusively on the factuality aspect of the news.
However, this aspect alone is insufficient to explain ``fake news,'' which is a complex phenomenon that involves a wide range of issues.
To fully understand the nature of each instance of fake news, it is important to observe it from various perspectives, such as the intention of the false news disseminator, the harmfulness of the news to our society, and the target of the news.
We propose a novel annotation scheme with fine-grained labeling based on detailed investigations of existing fake news datasets to capture these various aspects of fake news.
Using the annotation scheme, we construct and publish the first Japanese fake news dataset.
The annotation scheme is expected to provide an in-depth understanding of fake news.
We plan to build datasets for both Japanese and other languages using our scheme.
Our Japanese dataset is published at \url{https://hkefka385.github.io/dataset/fakenews-japanese/}.
 \\ \newline  }

\begin{document}

\maketitleabstract

\section{Introduction}
Fake news has caused significant damage to various fields of society, such as the economy, politics, and health problems.
For example, during the 2016 U.S. presidential election, 529 different low-credibility statements~\cite{president1} were spread on Twitter. 
Moreover, 25\% of the news outlets that were linked from tweets, which were either fake or extremely biased in supporting Trump or Clinton, potentially influenced the election~\cite{president2}.
Recently, the COVID-19 pandemic in 2020 resulted in the spread of disinformation and harmful content in the rapid influx of information, such as the relationship between the COVID-19 vaccine and infertility~\cite{infertility}.
Fake news has become a significant crisis that threatens a wholesome society and the social media ecosystem. 
{\let\thefootnote\relax\footnote{{* equal contribution}}}

Previous studies have proposed various tasks to combat the social problems caused by the spread of fake news.
For example, the fake news detection task aims to classify whether the news content that is spread from news articles and social media posts is false.
Additionally, many fake news datasets have been constructed as resources to facilitate the task, e.g., FakeNewsNet~\cite{shu2020fakenewsnet}, Twitter16~\cite{ma2017detect}, and CoAID~\cite{cui2020coaid}.
These existing studies on fake news detection and the corresponding dataset construction have focused almost exclusively on the factuality aspect of the news -- \textbf{Can we fully understand ``fake news'' and various events it causes based on these datasets given factuality labels?}
This is exactly the motivation behind our work.
To promote understanding of fake news, we consider that it is necessary to provide not only factual information, but also information from various perspectives, such as the intention of the false news disseminator, the harmfulness of the news to our society, and the target of the news.

We propose a novel annotation scheme to capture the various perspectives of false news, and it is based on our investigations of the definition of ``fake news'' and existing fake news detection datasets.
We annotate each news story and its social media posts using the following points: (1) factuality, (2) intention of the disseminator, (3) target, (4) method to report the target, (5) purpose, (6) potential harm to society, and (7) types of harm.
These annotations from various perspectives are useful in facilitating an in-depth understanding of fake news, which is a complex phenomenon. 
For example, it is interesting to consider how its spreading changes depending on the disseminator knowing whether the news is false or not.
The annotations also provide a significant value to real-world applications, such as building a fake news detection system that reveals the potential dangers of false information for journalists, fact-checkers, policymakers, and government entities.

We then construct a first Japanese fake news dataset according to the annotation scheme.
The construction of this dataset will facilitate our understanding of the spread of fake news in Japan.
In the future, we plan to apply this method to other fake news datasets in English and other languages.
Applying our annotation scheme to fake news in multiple countries and comparing the results is expected to enable a further detailed analysis of fake news.

This study makes the following contributions:
\begin{itemize}
    \item We identify issues that need to be resolved in dataset construction through a comprehensive survey of existing fake news detection datasets.
    \item We propose a novel annotation scheme to capture the news from various perspectives, instead of only considering factuality.
    \item We construct the first Japanese fake news dataset based on the annotation scheme.
\end{itemize}

\section{Survey}
First, we discuss the definition of ``fake news'' and its ambiguity.
We then identify issues that need to be resolved in the dataset construction through an exhaustive survey of existing fake news detection datasets.
Our proposed annotation scheme builds on the discussions and findings in this section.

\subsection{Definition of Fake News}
Researchers primarily employ either broad or narrow definitions of fake news.

A broad definition of fake news is that ``Fake news is false news.~\cite{zhou2020survey}''
Similarly, \newcite{lazer2018science} states that ``fake news is fabricated information that mimics news media content in form but not in organizational process or intent.''
This broad definition emphasizes only information authenticity and does not consider information intention.
This enables us to include different types of fake news, which can be identified by their motive or intent, such as satire and parody~\cite{rubin2015deception}.
A few studies~\cite{science2018,jin2016news} leverage the broad definition of fake news.

As the narrow definition of fake news, most research emphasizes its ``intention.''
\newcite{shu2017fake} and \newcite{allcott2017social} define fake news as ``a news article that is \textbf{intentionally} and verifiably false.''
\newcite{zhang2020overview} states that ``fake news refers to all kinds of false stories or news that are mainly published and distributed on the Internet, in order to \textbf{purposely} mislead, befool, or lure readers for financial, political, or other gains.''
Many other studies ~\cite{mustafaraj2017fake,conroy2015automatic,potthast2017stylometric} have also emphasized intention in the definition of fake news.
They adopt a narrow definition of fake news; nevertheless, their dataset construction only focus on the factuality of the news based on the judgment of fact-checking sites, not on the intention.

Therefore, the definition of the phrase ``fake news'' is ambiguous, and there is some criticism of this ambiguity.
For example, the British government decided that the phrase ``fake news'' would no longer be used in official documents because it is a poorly defined and misleading term that conflates a variety of false information~\cite{newsweek}.
Claire Wardle, the co-founder and leader of First Draft, announced that the phrase ``fake news'' is woefully inadequate to describe the related issues and distinguishes between three types of information-content problems: misinformation, disinformation, and malinformation~\cite{cnn_clair}.
Disinformation is related to the intention of users who create and share content, whereas malinformation is associated with the harmfulness of the information to society.
Part of our annotation scheme refers to this suggestion.
The concept of fake news has a variety of meanings, owing to current diverse circumstances.

\subsection{Issues in existing fake news detection datasets}
Many datasets have been constructed for the task of fake news detection, which assesses the truthfulness of a particular piece of news from news content or social media posts.
We examined 51 fake news detection datasets and identified four issues that needed to be resolved.
The details of each dataset are listed in  \newcite{murayama2021dataset}.


\begin{description}
    \item[Intention] Even though many studies adopt a narrow definition of fake news, which considers the intent of the disseminators, all datasets have labels that focus only on the factual aspects of each news item, not on the intention, based on the broad definition.
    This situation implies a divergence between the definition of fake news in technological development and the original narrow definition of fake news.
    We consider that most fake news detection models that are built on existing datasets should be called ``false information detection models.''
    Additionally, news created with malicious intent aims to be more persuasive than that without such aims, and malicious users typically participate in the propagation of false news to enhance its visibility on social media~\cite{leibenstein1950bandwagon}.
    This background makes it necessary to annotate the intentions of news disseminators to build a highly explainable detection model.

    \item[Harmfulness to society] 
    Fake news may have a greater or lesser detrimental effect on society.
    For example, parody news that is clearly false is less harmful to society; however, false news about elections or COVID-19 vaccines is very harmful, owing to its strong influence on people's decision-making.
    This perspective is not reflected in most existing datasets.
    A dataset called COVID-Alam~\cite{alam2021fighting} annotates each COVID-19 fake news item with its degree of harm to society.
    We consider that it would be useful for the decision on the priority of fact-checking to make a detailed annotation to various types of news, not only COVID-19 news.
    For example, it is important to consider which aspects of society the news is harmful to and what the extent of the harm is.
    This covers the malinformation perspective mentioned by Claire Wardle.

    \item[Languages]
    The linguistic characteristics and diffusion patterns of fake news vary according to country and language.
    However, the language included in most fake news datasets is English, and they primarily focus on the US society.
    This has occurred because although there is a growing awareness that fact-checking is an important action worldwide, there are still only a few fact-checking organizations with adequate workforce, which forms the basis for dataset construction, in countries other than the US.
    However, fake news detection datasets in languages other than English have also increased since the global infodemic caused by COVID-19.
    Of the 51 datasets that we examined, 11 included languages other than English, and 8 of them were datasets on COVID-19.
    The construction of a non-English fake news dataset that targets various topics leads to an analysis of fake news across languages and the identification of unique non-variant characteristics that are independent of language.

    \item[Labels] 
    33 datasets out of 51 are assigned a binary label, fake or real, because binary classification makes machine learning models easier to apply.
    Other datasets have fine-grained labels, typically more than two labels; however, the criteria vary across datasets based on the rating given by fact-checking sites; e.g., Politifact has six labels (True, Mostly True, Half True, Mostly False, False, and Pants on Fire) and Snopes primarily has five labels (True, Mostly True, Mixture, Mostly False, and False.)
    Such variation in the categorization criterion in each dataset confuses the dataset users.
    Further related to the above-mentioned issues of ``intention’’ and ``harmfulness to society,'' a fine-grained and consistent annotation scheme is required to build a more general and robust fake news detection model. 
\end{description}

\section{Annotation Scheme}
We present an annotation scheme that was developed through careful discussion and insights gained from an examination of existing datasets.
This section describes the key questions in our annotation.
Q1--Q5 are aimed at constructing a more fine-grained labeling than the binary labeling in existing datasets or the rating given by fact-checking sites.
These questions primarily cover ``intention'' and ``labels'' issues in existing datasets.
Q6 and Q7, which are extensions of \newcite{alam2021fighting} applied to general news, try to identify the harmful effects on society related to the second issue.
We ask annotators to answer these questions based on fact-checking articles and original texts.
The reclassification of false news using the shared annotation scheme with fine-grained labeling can achieve a common framework for understanding false news, which is independent of the rating of various fact-checking sites.
This is also useful in building detection models that are highly interpretive.

\noindent \textbf{Q1: What rating does the fact-checking site assign to the news?}
This is a very simple question, and it can be answered by simply searching the corresponding fact-checking site.
This also plays a role in removing inappropriate annotators.
Annotators generally choose a rating in the range between true and false, and the options vary depending on the fact-checking site.
If the annotators select True or Half-True, it implies that they automatically skip subsequent questions (Q2--Q7) that are asked only about false news.

\noindent \textbf{Q2-1: Does the news disseminator know that the news is false?}
This question asks for a subjective judgment. 
It covers ``intention,'' which is one of the issues in existing fake news detection datasets.
We ask the annotators to determine whether the spreading of fake news is intentional and classify their responses into four categories based on their observations of fact-checking articles and original social media posts.
If they select yes, the news can be considered fake news following the narrow definition; note that we call them ``disinformation'' based on \newcite{zhou2020survey}.
However, we cannot definitively regard the news as the news with malicious because it may be satire or parody news.
If they select no, it means that the disseminator does not intend to spread false news, which we call ``misinformation.''
Moreover, these decision branches are differentiated according to the degree of the annotator's belief, which is the distinction between ``definitely'' and ``probably.''
Such labeling of the intention may reveal the difference in users' behavior for each type of false information, such as the type of information people spread without knowing it is false.
This is also important irrespective of a study’s use of the broad or narrow definition.
The possible answers to Q2-1 are as follows:

\noindent 1. \textit{Yes, the news disseminator definitely knows that the news is false (Disinformation)}

\noindent 2. \textit{Yes, the news disseminator probably knows that the news is false (Disinformation)}

\noindent 3. \textit{No, the news disseminator probably does not know the news is false (Misinformation)}

\noindent 4. \textit{No, the news disseminator definitely does not know that the news is false (Misinformation)}

In addition, we set the following questions regarding the type of news, depending on the selection of Q2-1:

\noindent \textbf{Q2-2A: If yes (disinformation), how was the news created?}
This question is designed to annotate how intentionally disseminated news is created.
As a result of our detailed discussion and the analysis of a previous study~\cite{terms1}, we observed that each intentionally spread news story can be classified according to at least one of four categories: fabricated content, manipulated image, manipulated text, and false context.
First, these news stories can be categorized as either completely created news or news created by falsifying original resources.
We call the former ``fabricated content.''
The latter can be divided into three classes depending on the object of falsification: 
``manipulated image'' refers to content that has been manipulated for an image or video, ``manipulated text'' refers to content that has been manipulated for news text or social media messages related to the news, and ``false context'' refers to content that is shared with false contextual information despite the content itself being genuine.

\noindent 1. \textit{Fabricated content}

\noindent 2. \textit{Manipulated image}

\noindent 3. \textit{Manipulated text}

\noindent 4. \textit{False context}

\noindent \textbf{Q2-2B: If no (misinformation), how does the disseminator misunderstand the news?}
We label why the disseminator has spread the false news with no intention.
This is an important annotation for us to consider to prevent the future spread of false news.
Similar to Q2-2A, we observed that the reason for spreading false news with no intention can be classified into three categories: trusting other sources, inadequate understanding, and misleading.
The first category refers to trusting information from other sources.
This frequently occurs when non-native English speakers mistranslate English articles and research papers or trust false information that is originally disseminated in English.
The second category refers to an inadequate understanding and uncertain assumptions made by the disseminator.
This may be caused by the disseminator not having thoroughly read the news.
The final category refers to the case in which the disseminator may adequately understand the news, but they insufficiently convey it to the reader; that is, it refers to representing information in a misleading way.

\noindent 1. \textit{Trusting other sources}

\noindent 2. \textit{Inadequate understanding}

\noindent 3. \textit{Misleading}

\noindent \textbf{Q3: At whom or what is the false news targeted?}
The main target of false news, namely the target that is primarily affected by the fact that the news is false, is useful information for news clustering and retrieval.
The task of identifying such information has not yet been completed; however, we believe that it will be an important task to promote the understanding of fake news in the future.
To enable the application of information-extraction techniques, we give the annotators the following instructions: ``Extract the targets that are primarily affected by the fact that the news is false from the claim sentences on fact-checking sites or the original social media post about the false news (multiple extractions possible).''

\noindent \textbf{Q4: Does the news flatter or denigrate the target?}
We annotate the stance that the news has toward the target, that is, flattery or denigration.
Even within the category of false news, the reader’s impression of good behavior news, such as donations, is very different from that of bad behavior news, such as criminal acts, even though the target has not actually done either act. 
This annotation provides important information for understanding the impact of fake news on society, particularly for analyzing the impact of fake news on polarization. 
The annotations are as follows:

\noindent 1. \textit{Flattery}

\noindent 2. \textit{Denigration}

\noindent 3. \textit{Neither / No such intention}

\noindent \textbf{Q5: What is the purpose of the false news?}
Just as some news media lean toward liberal or conservative views and report the news according to it, some false news stories are fabricated with some intention to spread the disseminator’s own theory; for example, the COVID-19 vaccine is dangerous to human health.
Although the purpose of some false news items cannot be inferred, we set the following categories of false news purpose.
The first category is satire or parody news for the purpose of entertaining or criticizing readers~\cite{brummette2018read}.
These false news stories are not commonly referred to as fake news.
The second is partisan news, which is extremely one-sided or biased news that has a political context.
Biasing in itself does not mean that the news is fake; however, some studies~\cite{icsss2,zannettou2018gab} report that it has a high possibility of being false in parts of partisan news.
This annotation is important to understand the relationship between partisan news and false news.
The third is propaganda, which is a form of persuasion that attempts to influence the emotions, attitudes, opinions, and actions of specified target audiences for ideological, religious, and other purposes~\cite{jowett2018propaganda}.
Propaganda may also include political purposes in general; however, we instruct the annotators to categorize propaganda with political purposes in the partisan category to aid distinction.
This question is expected to clarify the relationship between false news and the following categories:

\noindent 1. \textit{Satire / Parody}

\noindent 2. \textit{Partisan}

\noindent 3. \textit{Propaganda}

\noindent 4. \textit{No purpose / Unknown}

\noindent \textbf{Q6: To what extent is the news harmful to society?}
This is a particularly subjective question.
Its purpose is to identify news stories that can negatively affect society, including specific people and companies.
Specifically, we ask the annotators to indicate the degree of harm to society on a real scale of 0--5.
A score of 0 indicates that the news poses no harm to society.
A score of 5 indicates that the news is definitely harmful to society.
To obtain the annotators’ answers, which do not vary greatly, we ask them to label the degree of harm using a combination of two perspectives: how much truth is in the text description, and how much damage may be caused by believing the news.

\noindent \textbf{Q7: What types of harm can the news cause?}
This question helps us understand what types of harm the news causes or has the potential to cause.
We set up seven categories of harm that fake news can cause, and we added the option ``not sure'' for cases in which a decision cannot be made.
The categories are described below.
Some news stories may be aligned with more than one category; however, we ask the annotators to choose one category that they consider the most appropriate.

\noindent 1. \textit{Harmless (e.g., Satire / Parody)}

\noindent 2. \textit{Confusion and anxiety about society}

\noindent 3. \textit{Threat to honor and trust in people and companies}

\noindent 4. \textit{Threat to correct understanding of politics and social events}

\noindent 5. \textit{Health}

\noindent 6. \textit{Prejudice against country and race}

\noindent 7. \textit{Conspiracy Theory}

\noindent 8. \textit{Not sure}

\begin{figure}[t]
    \centering
    \includegraphics[width=\linewidth]{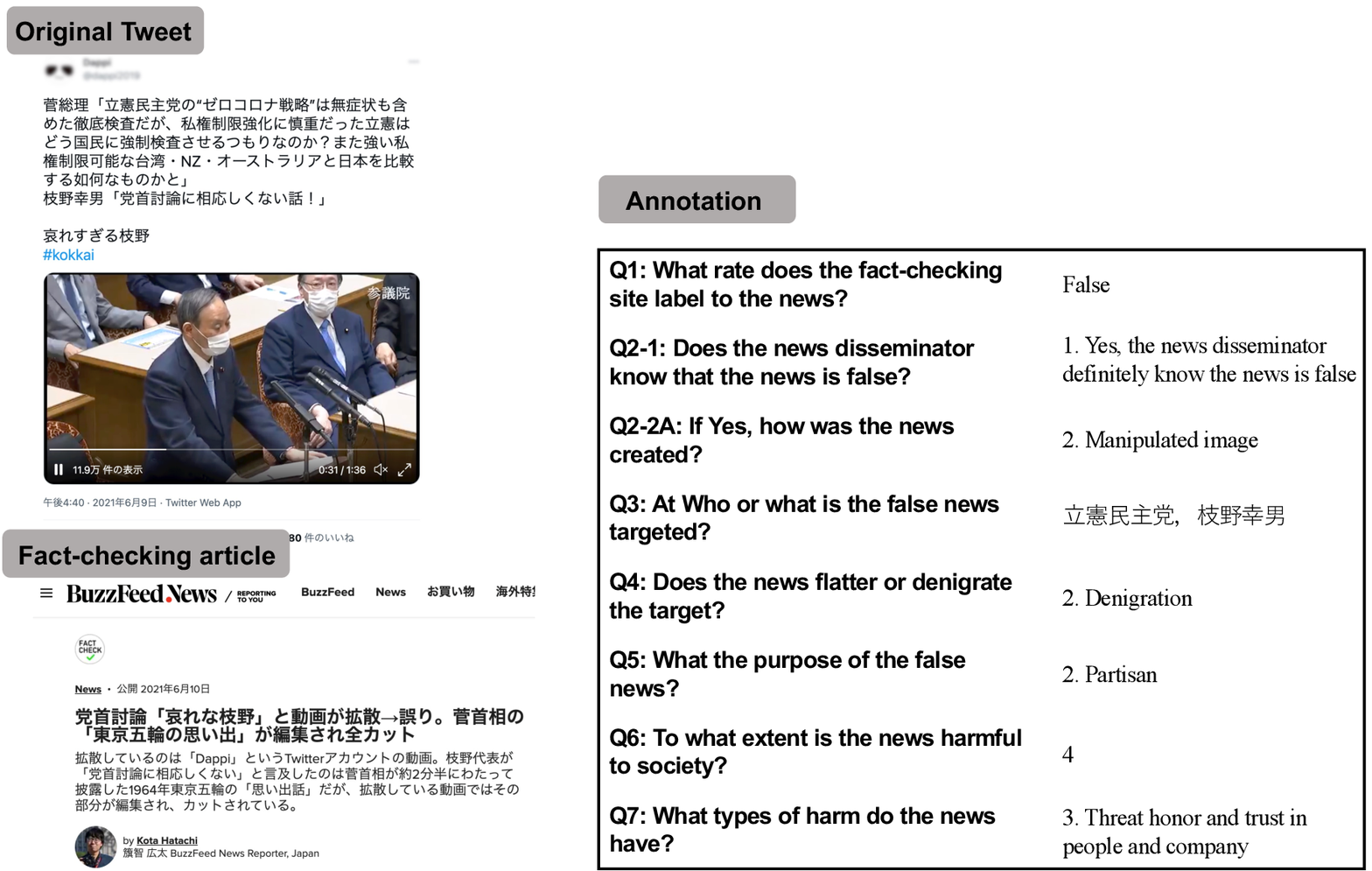}
    \caption{Original tweet and the corresponding fact-checking article of its targeting for our annotation are shown on the left. Labeled information by our annotation is shown on the right. The targeting content is a video of a party debate attached by a social media influencer on Twitter. 
    It is stated in the fact checker’s judgment that this video creates a bad impression of the opposition leader (Mr. Edano) because it omits parts of the debate.
    } 
    \label{example1}
\end{figure}

\begin{figure}[t]
    \centering
    \includegraphics[width=\linewidth]{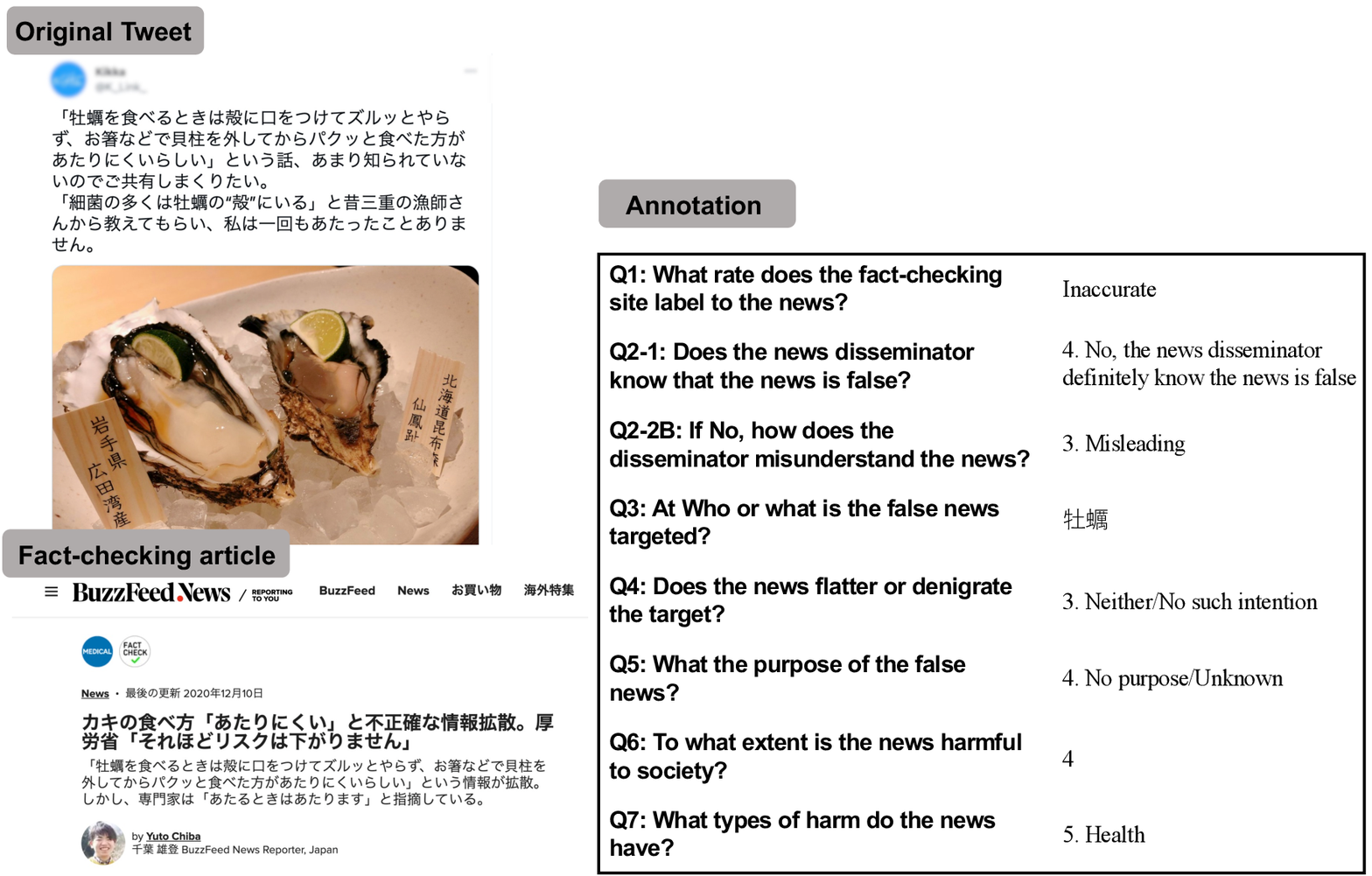}
    \caption{The targeting content describes how to eat oysters for the prevention of food poisoning. The fact checker notes that the method prescribed for eating does not reduce the likelihood of food poisoning.}
    \label{example2}
\end{figure}

\section{Japanese Fake News Dataset}
\subsection{Original Data}
To construct the Japanese fake news dataset following our annotation scheme, we first collected verified articles published in Fact Check Initiative Japan~\cite{FIJ}.
We targeted the news that was spread on Twitter via these verified articles.
And we manually searched for posts or news articles that triggered the spread of false information using Twitter search function.
We asked the annotators to annotate 307 news stories, which were featured by Fact Check Initiative Japan between July 2019 and October 2021.
Examples of these annotations are shown in Figures~\ref{example1} and~\ref{example2}.

\subsubsection{Annotation}
As a pilot annotation, four annotators independently annotated 20 examples and attempted to resolve cases of disagreement in a meeting.
Based on their discussions, the annotation scheme and guidelines were refined.
Finally, we asked three annotators to answer the questions introduced in the Annotation section regarding the 307 verified news stories by checking the verification articles, triggered posts, and news articles.
In the annotation process, we calculated the inter-annotator agreement using Fleiss’ kappa.
The results show that Fleiss’ Kappa was generally high for each question.
For example, it was higher than 0.8 for Q2-1 and 0.7 for Q4.
Additionally, it was higher than 0.6 for Q7, for which eight options exist.
In contrast, Fleiss’ kappa for Q2-2A and Q2-2B, which were subjective questions, was approximately 0.5.
Note that kappa values of 0.21--0.40, 0.41--0.60, 0.61--0.80, and 0.81--1.0, correspond to fair, moderate, substantial, and perfect agreement, respectively~\cite{kappa}.

\subsubsection{Data statistics}
Table~\ref{annotate_table} shows relevant statistics on the annotations.
Q1 shows the distribution of the fact-checking judgment for each news story.
Most articles selected by Japanese fact-checking organizations for verification are false stories.
Thus, the selection of articles by Japan's fact-checking organization is biased (only five news stories are true). 
For Q2-1, the labels ``disinformation'' and ``misinformation'' were applied to 13\% and 87\% of the news stories, respectively. 
In most cases, the disseminator was unaware that the news was false.
The class distribution of Q2-2A, for which only the news stories labeled as disinformation were annotated, is relatively balanced.
In Q2-2B, for misinformation, ``inadequate understanding'' accounts for approximately half of the annotations.
For Q4, which asks whether the news flatters or denigrates the target, the distribution is skewed towards ``denigration'' in 60\% of the news stories. 
This suggests that most false news is written to discredit people.
For Q5, which asks what the purpose of the false news is, most news stories are labeled as ``no purpose / Unknown.'' 
Propaganda and partisan false news were identified in approximately 20\% of news stories each.
For Q6, the extent to which the news is harmful to society, the annotators chose average scales of 1--2 and 2--3 for many false news stories from a range of 0--5.
For Q7, which asks what types of harm the news has, the majority of news stories are labeled as ``threat to honor and trust in people and companies'' (36\%).
Most news stories labeled as ``health'' are related to COVID-19.
The ``harmless'' and ``conspiracy theory'' labels only constitute a small percent of new stories.
Our fine-grained annotations can be a useful tool for understanding false news trends in the target country.

In addition to the annotation results, we collected posts and related context information on 186 news stories that triggered the spread of false information from Twitter using Twitter Search API.
The data we collected from Twitter included 471,446 tweets (2,534 tweets per news story), 277,106 users (1,489 users per news story), and 17,401 conversations (93 conversations per news story).
We publish these annotation results, the collected tweet IDs, fact-checked articles, and other related information in \url{https://hkefka385.github.io/dataset/fakenews-japanese/}.

\begin{table}[!t]
    \centering
    \scriptsize
    \caption{Distribution of the Japanese fake news dataset. In the rows with a question, we show the total number of annotations.}
    \begin{tabular}{l|r} \hline
         Q1: What rating does the fact-checking site attribute to the news? & 307 \\ \hline
         True & 1\\
         Half-True & 4\\
         Inaccurate & 50\\
         Misleading & 52 \\
         False & 153\\
         Pants on Fire & 16\\
         Unknown Evidence & 30\\
         Suspended Judgment & 1\\ \hline
         Q2-1: Does the news disseminator know that the news is false? & 301 \\ \hline
         1. Yes, the news disseminator definitely knows that & \multirow{2}{*}{20}\\
         the news is false. (Disinformation) & \\
         2. Yes, the news disseminator probably knows & \multirow{2}{*}{19}\\
         that the news is false. (Disinformation) & \\
         3. No, the news disseminator probably does not knows & \multirow{2}{*}{155}\\
         that the news is false. (Misinformation) & \\
         4. No, the news disseminator definitely does not knows & \multirow{2}{*}{107}\\ 
         that the news is false. (Misinformation) & \\ \hline
         Q2-2A: If yes, how was the news created? & 39\\ \hline
         1. Fabricated content & 15\\
         2. Manipulated image & 12\\
         3. Manipulated text & 6\\
         4. False context & 6 \\ \hline
         Q2-2B: If no, how does the disseminator misunderstand the news? & 262\\ \hline
         1. Trusting other sources & 61\\
         2. Inadequate understanding & 131\\
         3. Misleading & 70\\ \hline
         Q4: Does the news flatter or denigrate the target? & 301\\ \hline
         1. Flattery & 25\\
         2. Denigration & 181\\
         3. Neither / No such intention & 95\\ \hline
         Q5: What is the purpose of the false news? & 301\\ \hline
         1. Satire / Parody & 6\\
         2. Partisan & 70\\
         3. Propaganda & 67\\ 
         4. No purpose / Unknown & 158 \\\hline
         Q6: To what extent is the news harmful to society? (average) & 301\\ \hline
         0 $\sim$ 1 (including 1) & 17\\
         1 $\sim$ 2 (including 2) & 128\\
         2 $\sim$ 3 (including 3) & 112\\
         3 $\sim$ 4 (including 4) & 41\\
         4 $\sim$ 5 (including 5) & 3\\
         \hline
         Q7: What types of harm can the news cause? & 301 \\ \hline
         1. Harmless (e.g., Satire / Parody) &  6\\
         2. Confusion and anxiety about society & 41\\
         3. Threat to honor and trust in people and companies & 109\\
         4. Threat to correct understanding of politics and social events & 63\\
         5. Health & 29\\
         6. Prejudice against country and race & 42\\
         7. Conspiracy theory & 11\\
         8. Not sure & 0\\ \hline
    \end{tabular}
    \label{annotate_table}
\end{table}


\section{Analysis of the Japanese Fake news dataset}
Our dataset includes multi-dimensional information related to news content and social context.
We provide some preliminary quantitative analyses to illustrate the characteristics of the dataset.
In general, we analyze news stories from the perspectives of true and fake. 
However, our dataset includes few news stories labeled as true; therefore, this section primarily focuses on the news stories from the perspectives of misinformation and disinformation obtained from Q2-1.
Note that the analysis does not cover most fake news in Japan, but only news verified by fact-checking organizations.
Thus, a bias in the news stories may exist.

\subsection{Tweet contents}
\begin{CJK}{UTF8}{ipxm}
We aim to understand what news story topics spread in each category.
Therefore, we created a word cloud from the contents of each tweet that spreads most extensively on Twitter according to each news story.

\begin{figure}[t!]
\begin{minipage}[b]{0.48\linewidth}
    \centering
    \captionsetup{width=.90\linewidth}
    \includegraphics[width=\textwidth]{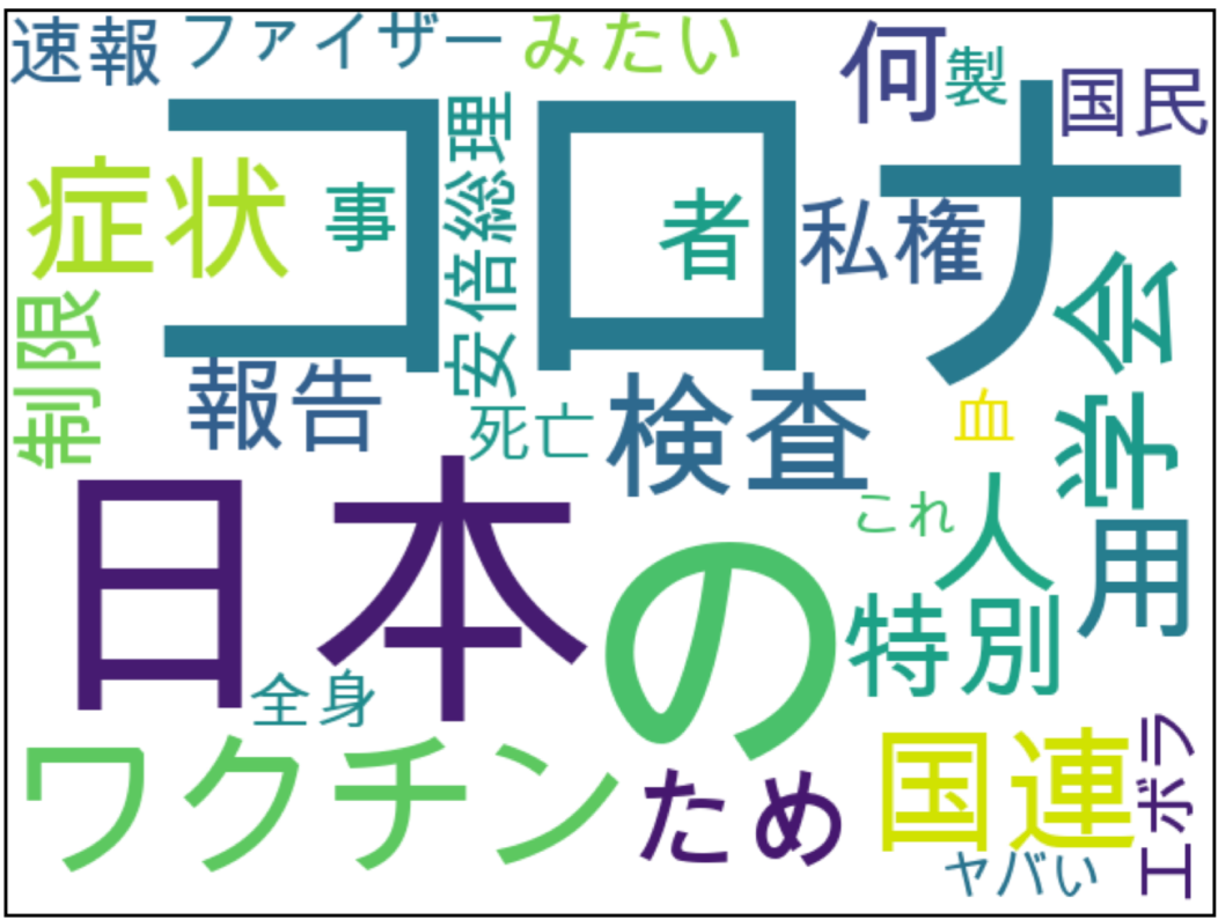}
    \subcaption{Disinformation (Labeled 1 or 2)}
\end{minipage}
\begin{minipage}[b]{0.48\linewidth}
    \centering
    \captionsetup{width=.90\linewidth}
     \includegraphics[width=\textwidth]{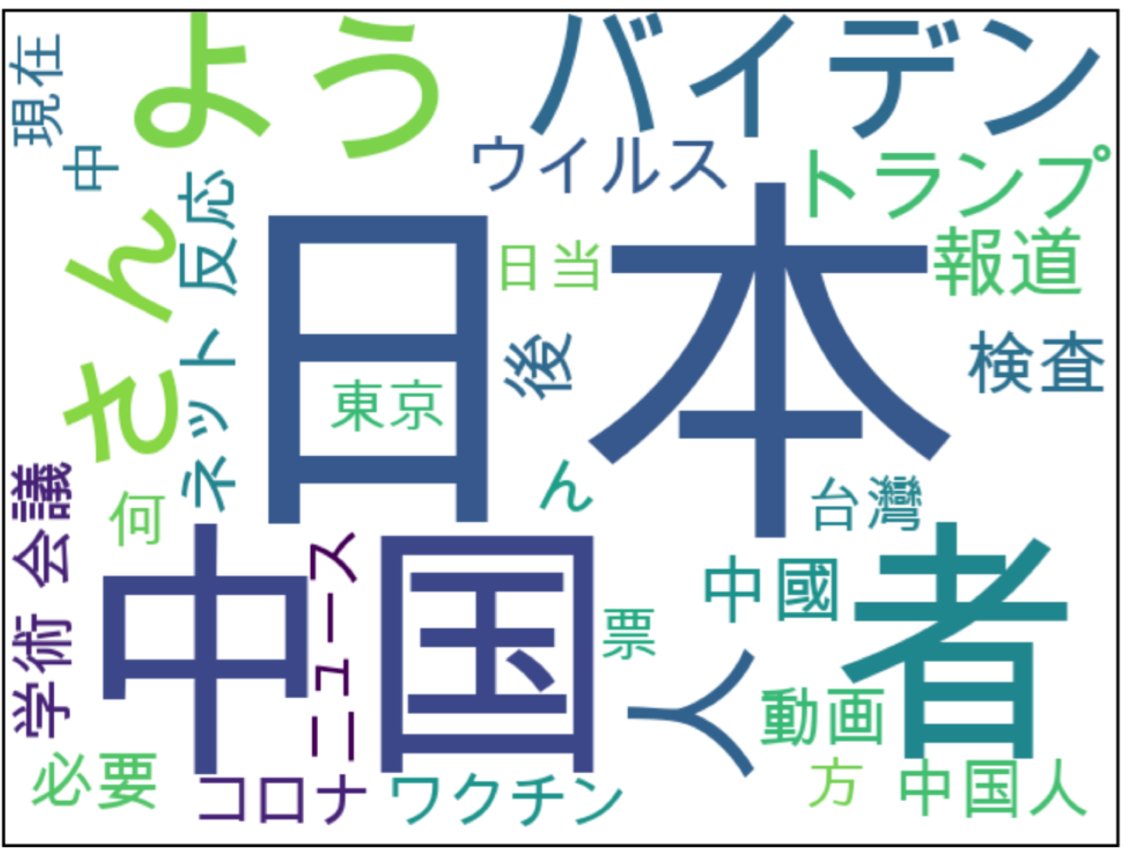}
    \subcaption{Misinformation (Labeled 3 or 4)}
\end{minipage}
\caption{Word cloud for ``Q2-1: Does the news disseminator know that the news is false?''}
\label{wordcloud1}
\end{figure}

From Figure~\ref{wordcloud1}, we can observe the contents of the disinformation and misinformation in the news stories based on the labels used for Q2-1.
The word ``日本 (Japan)'' is prominent in both word clouds.
In particular, in Figure~\ref{wordcloud1}(a), the word cloud for news stories labeled disinformation contains the words ``コロナ (coronavirus)'' and ``ワクチン (vaccine),'' which are related to COVID-19.
However, the word cloud for news stories labeled misinformation in Figure~\ref{wordcloud1}(b) contains the words ``中国 (China),'' ``バイデン (Mr. Biden),'' and ``トランプ (Mr. Trump),'' which are related to names of foreign people and countries.

\begin{figure}[t]
\begin{tabular}{c}
\begin{minipage}[t]{0.30\linewidth}
    \centering
    \captionsetup{width=.90\linewidth}
    \includegraphics[width=\textwidth]{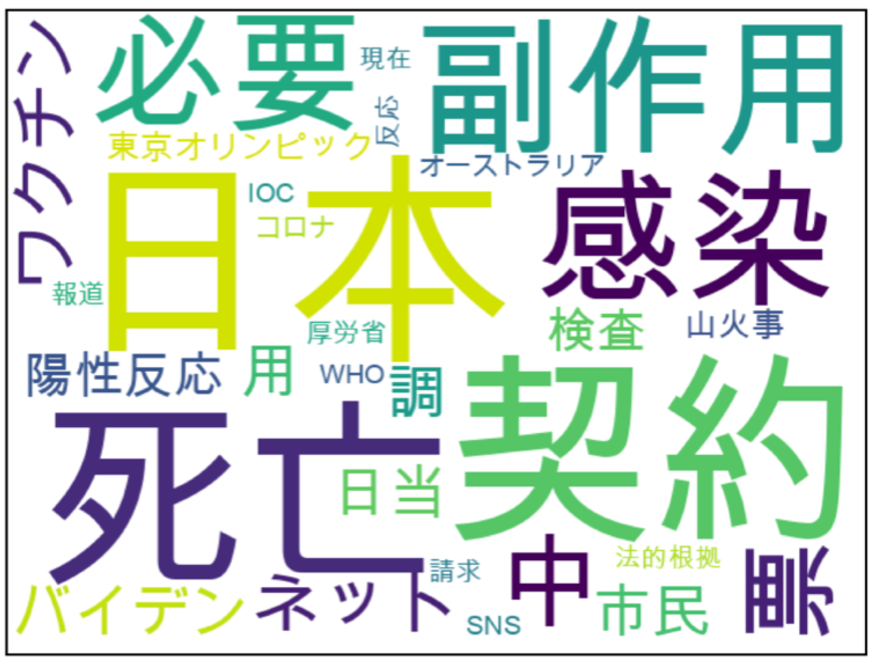}
    \subcaption{2. Confusion and anxiety about society}
\end{minipage}
\begin{minipage}[t]{0.30\linewidth}
    \centering
    \captionsetup{width=.90\linewidth}
    \includegraphics[width=\textwidth]{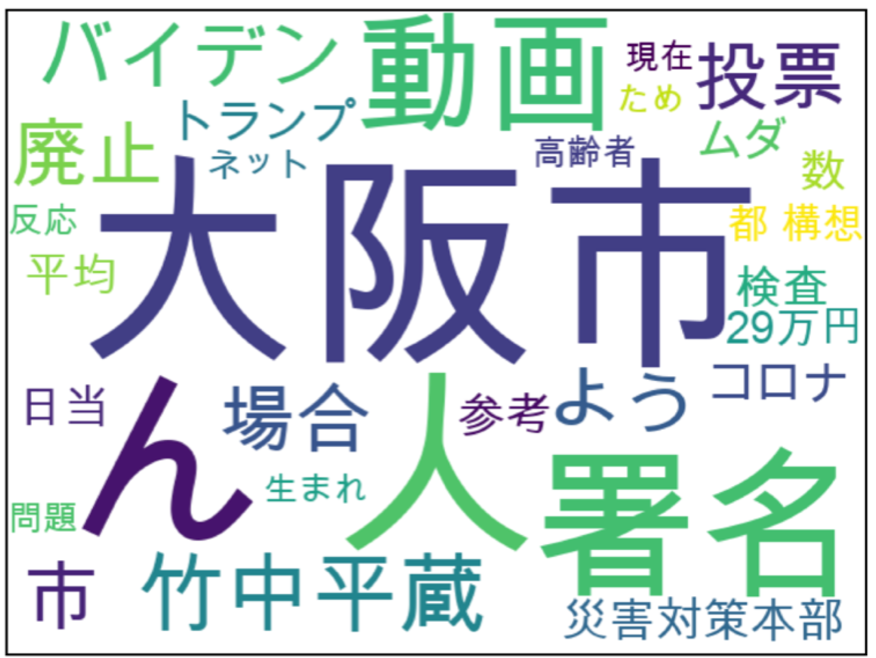}
    \subcaption{3. Threat to honor and trust in people and companies}
\end{minipage}
\begin{minipage}[t]{0.30\linewidth}
    \centering
    \captionsetup{width=.90\linewidth}
     \includegraphics[width=\textwidth]{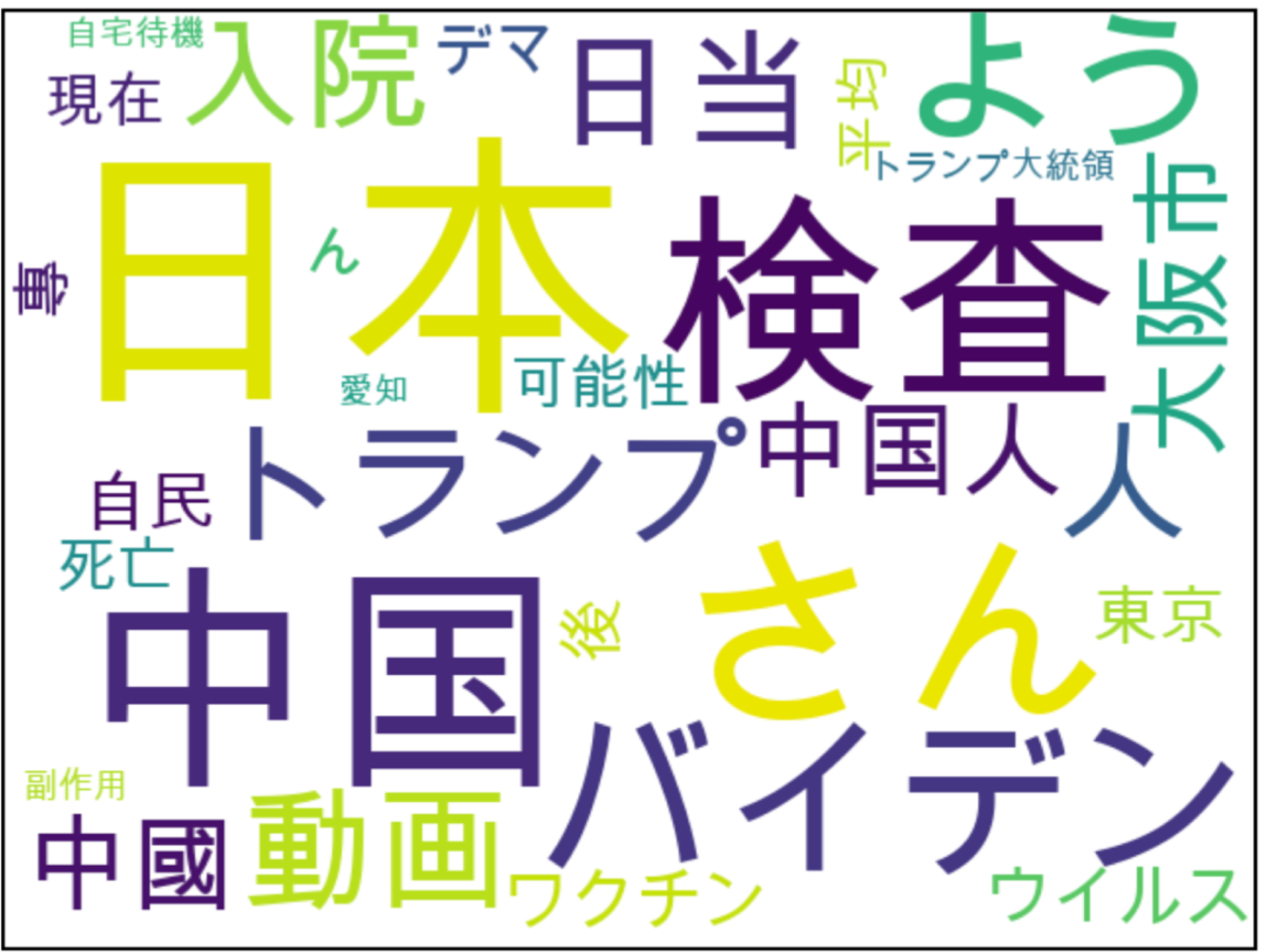}
    \subcaption{4. Threat to the correct understanding of politics and social events}
\end{minipage}\\
\vspace{5pt}
\begin{minipage}[t]{0.30\linewidth}
    \centering
    \captionsetup{width=.90\linewidth}
    \includegraphics[width=\textwidth]{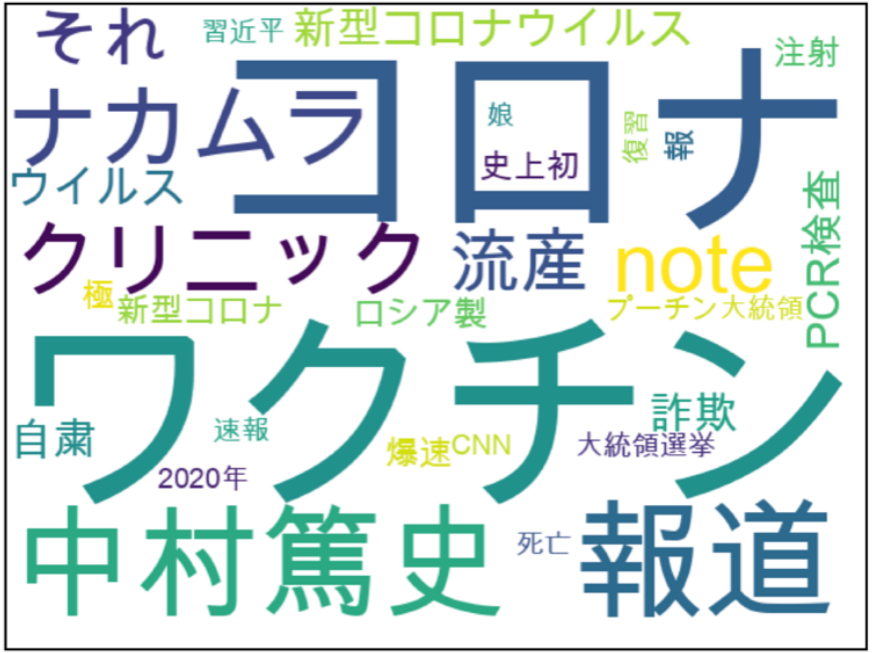}
    \subcaption{5. Health}
\end{minipage}
\begin{minipage}[t]{0.30\linewidth}
    \centering
    \captionsetup{width=.90\linewidth}
    \includegraphics[width=\textwidth]{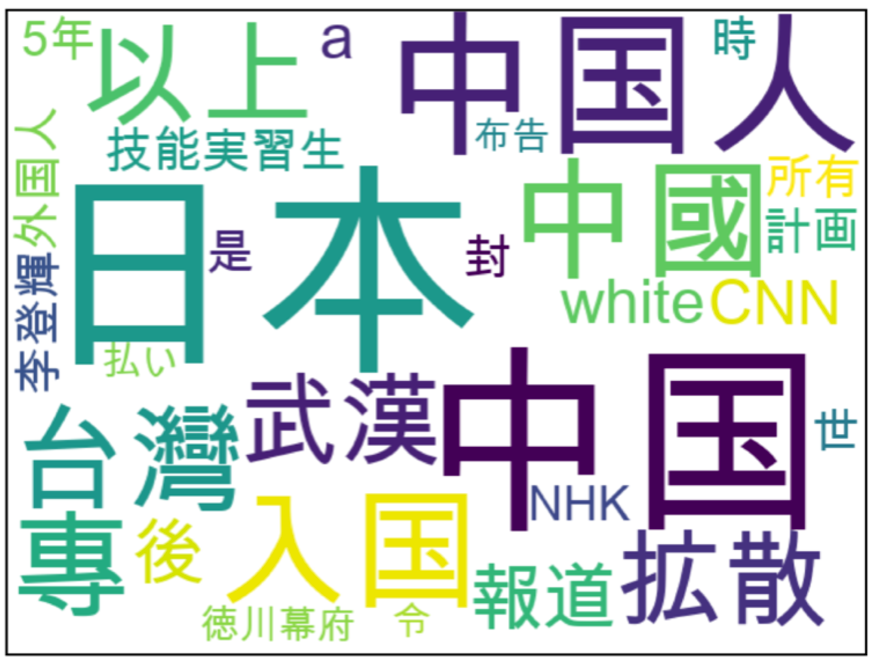}
    \subcaption{6. Prejudice against country and race}
\end{minipage}
\begin{minipage}[t]{0.30\linewidth}
    \centering
    \captionsetup{width=.90\linewidth}
     \includegraphics[width=\textwidth]{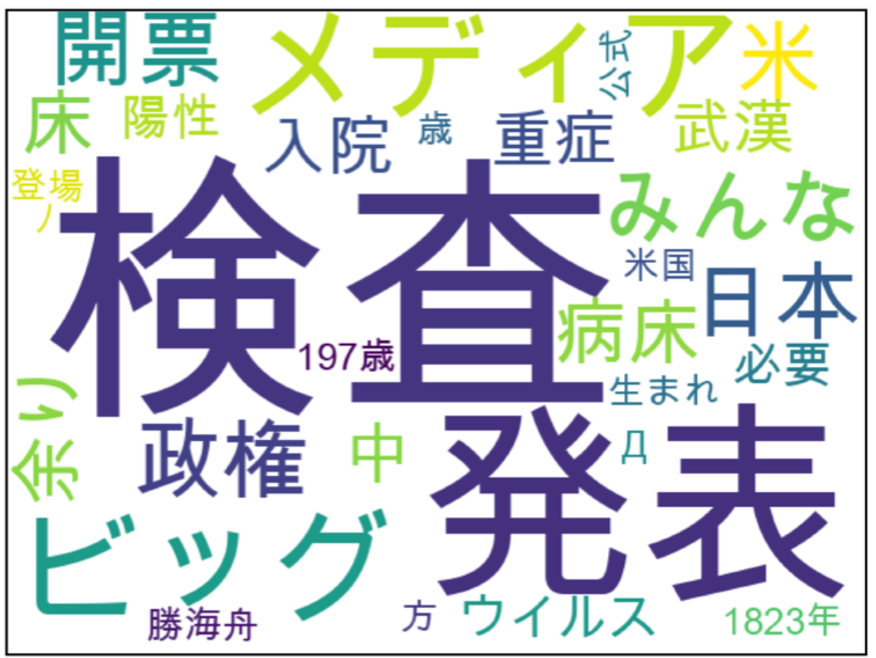}
    \subcaption{7. Conspiracy theory}
\end{minipage}\\
\end{tabular}
\caption{Word cloud for ``Q7: What types of harm can the news cause?''}
\label{wordcloud4}
\end{figure}

Figure~\ref{wordcloud4} shows the word clouds for news stories based on Q7, which asks what types of harm the news can cause.
Figure~\ref{wordcloud4}(a), which exhibits a word cloud on news stories labeled as confusion and anxiety about society, has the words ``感染 (infection),'' ``死亡 (death),'' and ``副作用 (side effects),'' which describe anxiety about COVID-19 and its vaccines.
Figure~\ref{wordcloud4}(b), which displays a word cloud on news stories labeled as a threat to honor and trust in people and companies, includes the words ``大阪市 (Osaka city, a regional city in Japan).''
This suggests that false news stories about local elections and the government in Osaka can be attributed to this category.
Figure~\ref{wordcloud4}(c), which shows a word cloud on news stories labeled as a threat to the correct understanding of politics and social events, includes the words ``中国 (China),'' ``バイデン (Mr. Biden),'' and ``トランプ (Mr. Trump).''
This indicates that news stories that promote a false understanding of foreign events in China and the US are being spread.
Figure~\ref{wordcloud4}(d), which displays a word cloud on news stories labeled health, includes the words ``コロナ (coronavirus),'' ``ワクチン (vaccine),'' and ``クリニック (clinic),'' which are related to COVID-19 events.
Figure~\ref{wordcloud4}(e), which exhibits a word cloud on news stories labeled as prejudice against country and race, has the words ``中国 (China),'' ``中國 (China in Chinese),'' and ``中国人 (Chinese person).''
This suggests that there is a lot of false news that may cause prejudice against China.
Figure~\ref{wordcloud4}(f), which shows a word cloud on news stories labeled conspiracy theory, includes the words ``ビッグ (big)'' and ``発表 (announcement).''
It seems that these words are often used when people want to spread conspiracy theories.
\end{CJK}


\captionsetup[figure]{font=small}
\captionsetup[figure]{font=normalsize}

\subsection{Sentiment of responses}
People express their emotions or opinions about false news through social media posts, such as skeptical opinions and sensational reactions. 
These features are important signals for the study of false news in general~\cite{qazvinian2011rumor,jin2016news}.

\begin{figure}[t]
\begin{minipage}[t]{0.45\linewidth}
    \centering
    \captionsetup{width=.90\linewidth}
    \includegraphics[width=\textwidth]{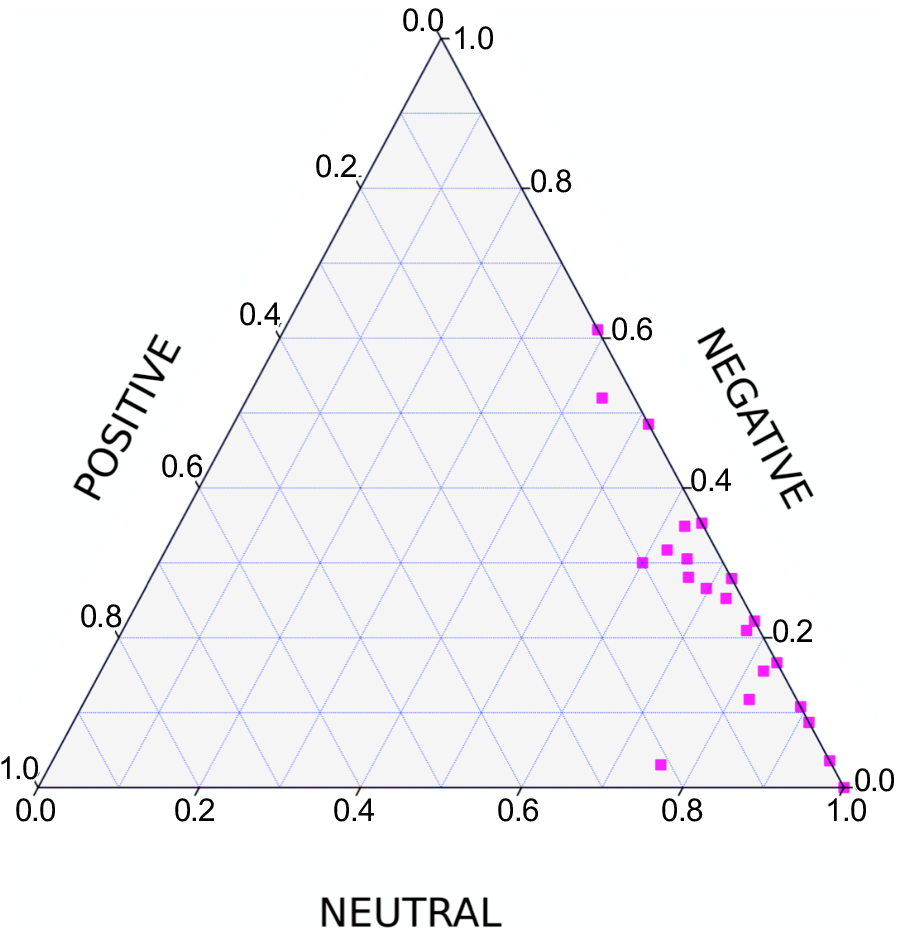}
    \subcaption{Sentiment about disinformation (Labeled 1 or 2)}
\end{minipage}
\begin{minipage}[t]{0.45\linewidth}
    \centering
    \captionsetup{width=.90\linewidth}
     \includegraphics[width=\textwidth]{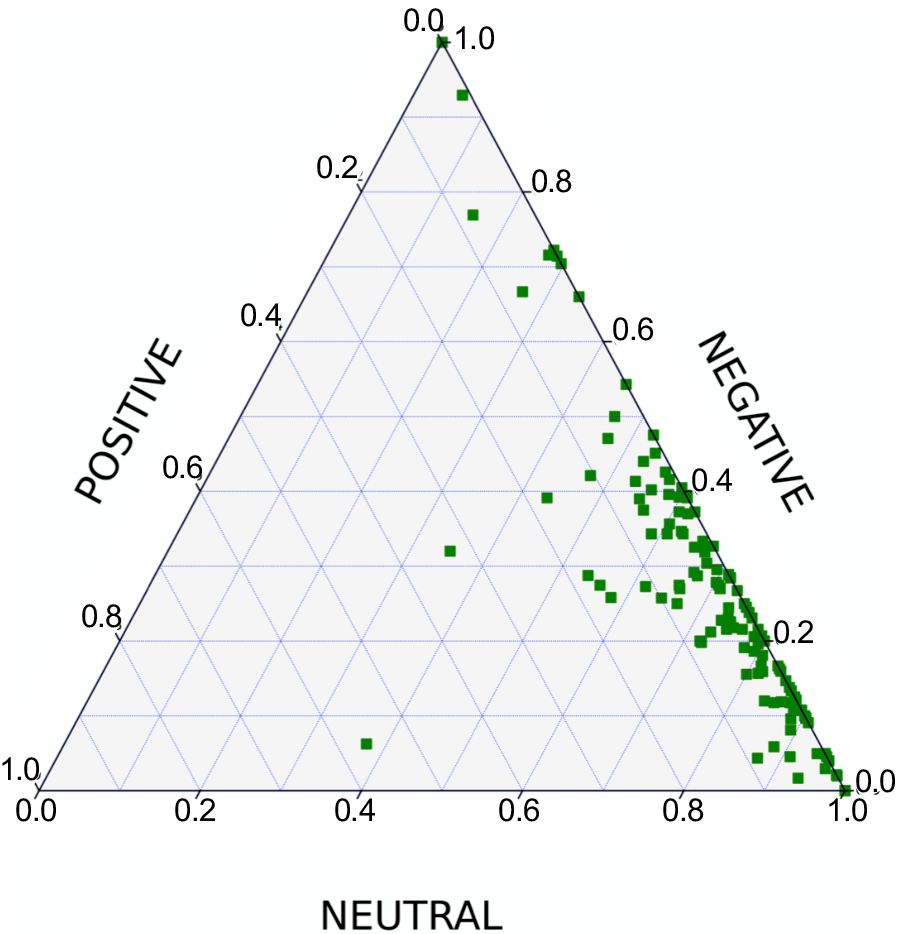}
    \subcaption{Sentiment about misinformation (Labeled 3 or 4)}
\end{minipage}
\caption{Ternary plot of the ratio of the positive, neutral, and
negative sentiment replies to tweets related to news labeled as disinformation and misinformation in ``Q2-1: Does the news disseminator know that the news is false?''}
\label{posneg}
\end{figure}

We performed sentiment analysis on the replies to user posts that spread false news using the sentiment classification API in Amazon Comprehend~\cite{ams}, which leverages a pretraining language model.
This API classifies emotions from the input text into one of four categories: positive, negative, neutral, or mixed.
Figure~\ref{posneg} shows the relationship between the positive, neutral, and negative replies to news stories from the perspectives of disinformation and misinformation obtained from Q2-1.
It represents the ratio of sentiments (positive, negative, or neutral), which are predicted from all the replies to the related tweets of each news story.
The ternary plots of both disinformation and misinformation show that most replies to each news item are neutral instead of emotional responses.
An analysis of the emotional replies shows that although some news stories that were labeled misinformation had a high ratio of positive replies, most news stories had more negative replies.
It is suggested that false news is likely to cause negative emotions, regardless of misinformation and disinformation.

\begin{figure}[t]
\begin{tabular}{c}
\begin{minipage}[t]{0.45\linewidth}
    \centering
    \captionsetup{width=.90\linewidth}
    \includegraphics[width=\textwidth]{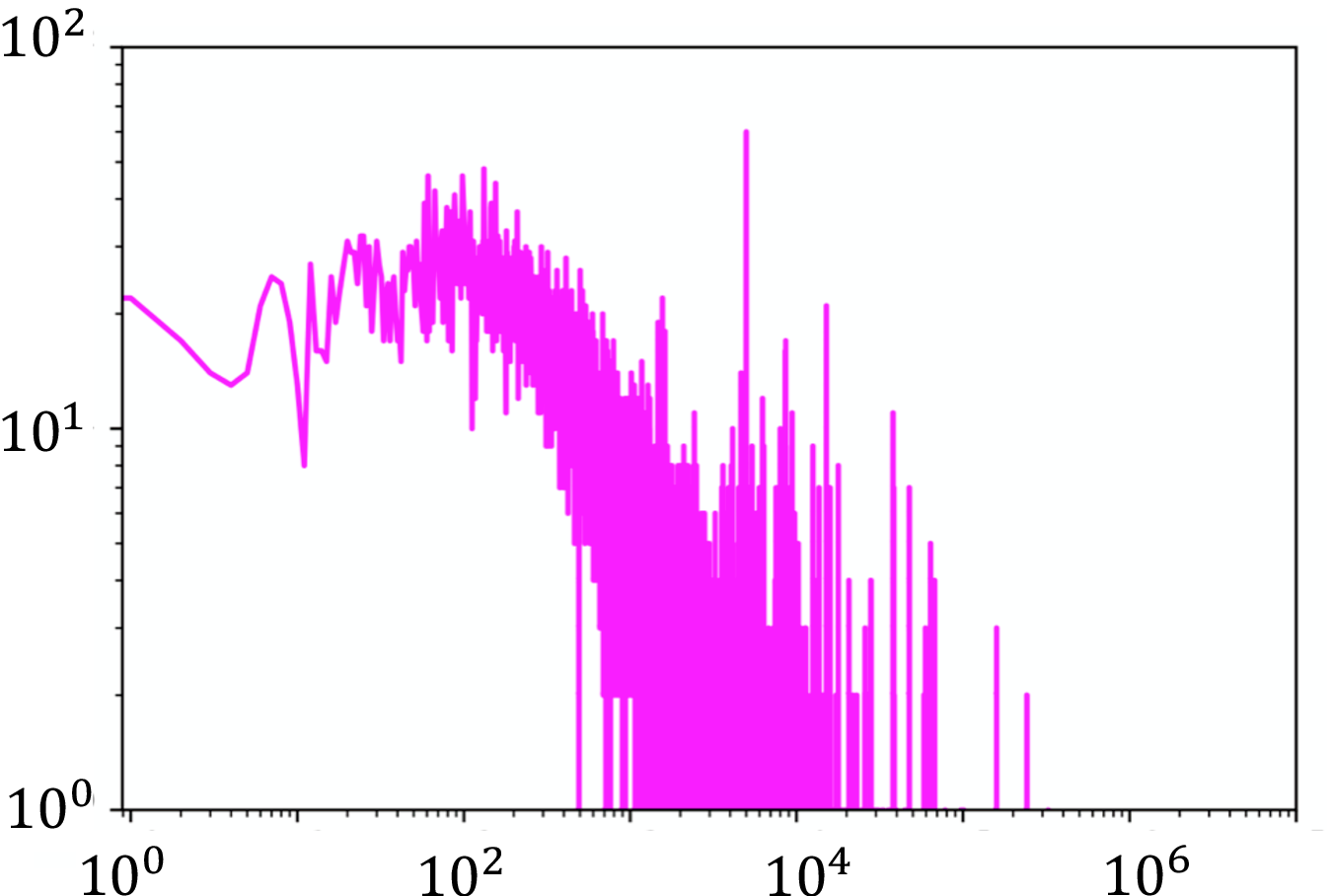}
    \subcaption{Followee count of users who posted tweets labeled as disinformation (Labeled 1 or 2)}
\end{minipage}
\begin{minipage}[t]{0.45\linewidth}
    \centering
    \captionsetup{width=.90\linewidth}
     \includegraphics[width=\textwidth]{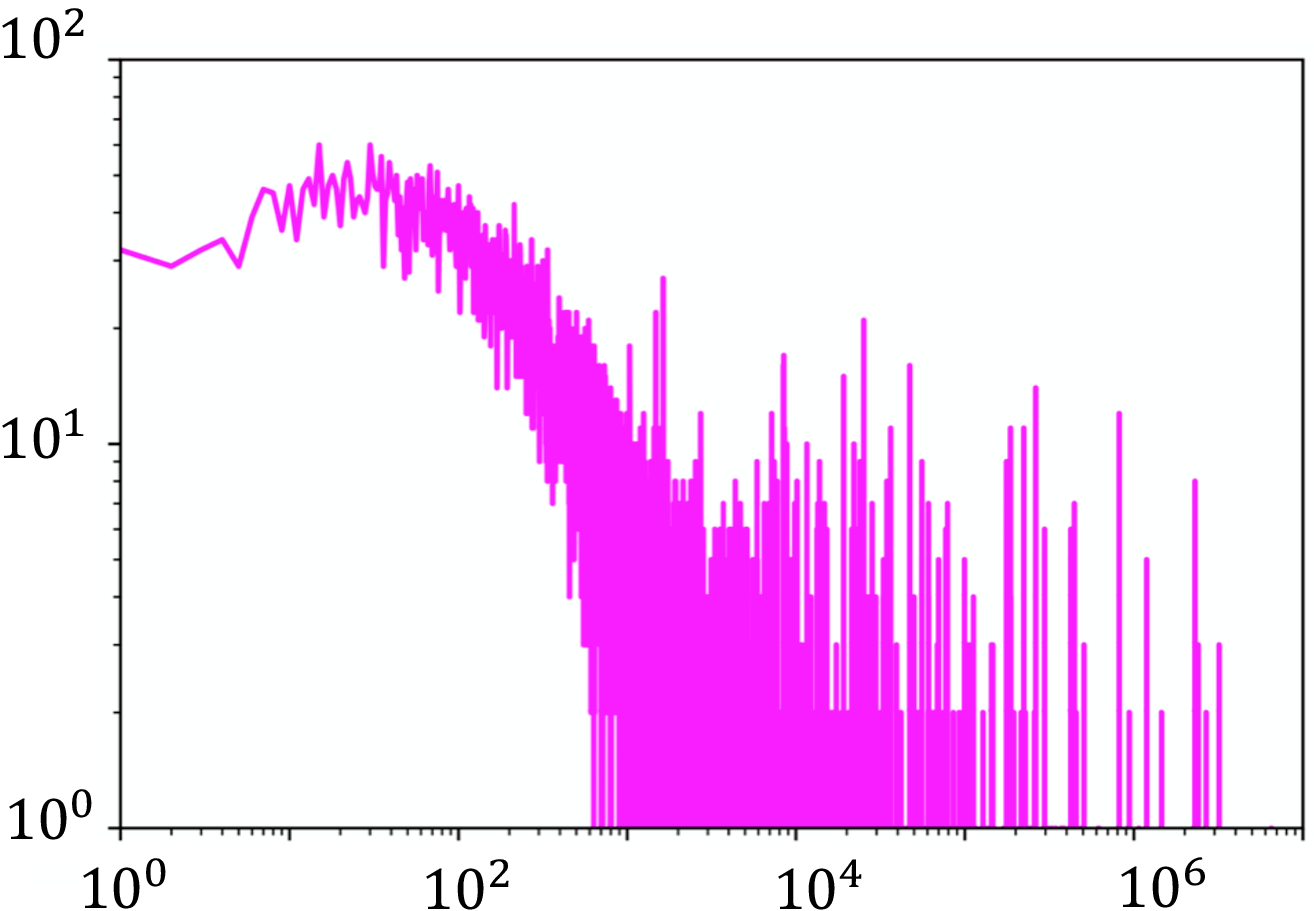}
    \subcaption{Follower count of users who posted tweets labeled as disinformation (Labeled 1 or 2)}
\end{minipage} \\
\begin{minipage}[t]{0.45\linewidth}
    \centering
    \captionsetup{width=.90\linewidth}
    \includegraphics[width=\textwidth]{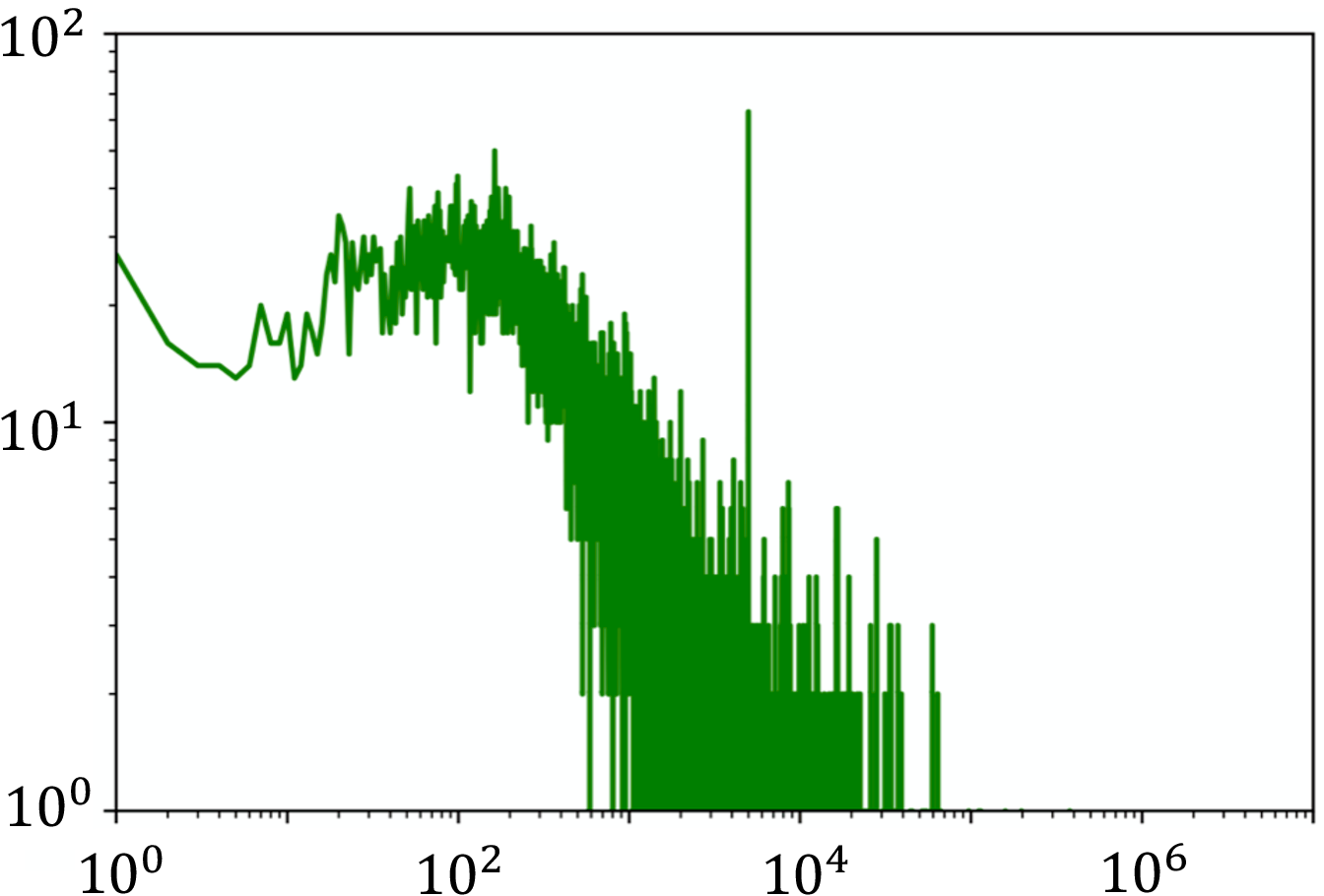}
    \subcaption{Followee count of users who posted tweets labeled as misinformation (Labeled 3 or 4)}
\end{minipage}
\begin{minipage}[t]{0.45\linewidth}
    \centering
    \captionsetup{width=.90\linewidth}
     \includegraphics[width=\textwidth]{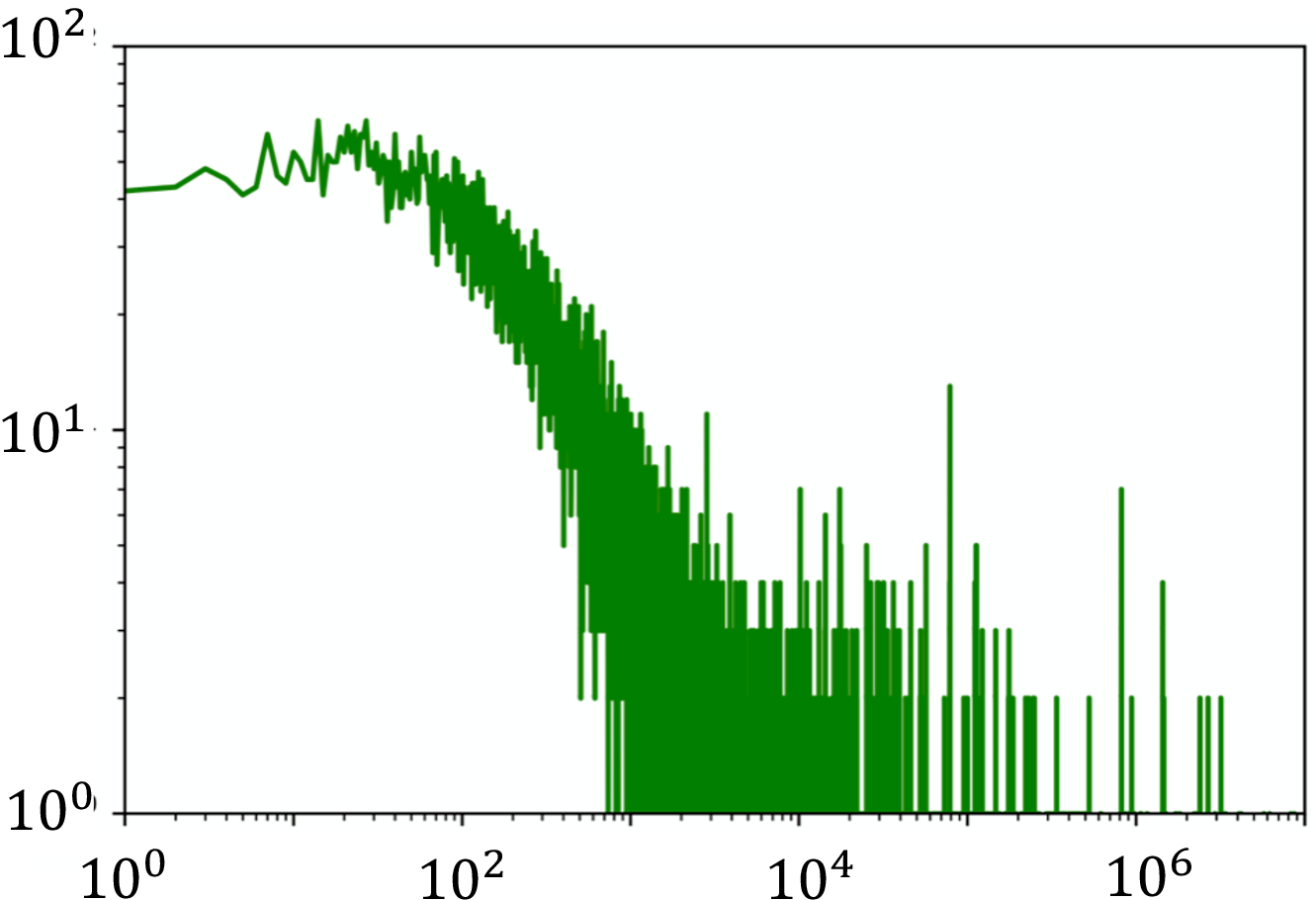}
    \subcaption{Follower count of users who posted tweets labeled as misinformation (Labeled 3 or 4)}
\end{minipage} \\

\end{tabular}
\caption{The distribution of the follower and followee count related to tweets labeled as disinformation or misinformation. X-axis represents the follower/folowee count and Y-axis represents the number of users.}
\label{follower}
\end{figure}

\begin{figure}[t!]
    \centering
    \includegraphics[width=\linewidth]{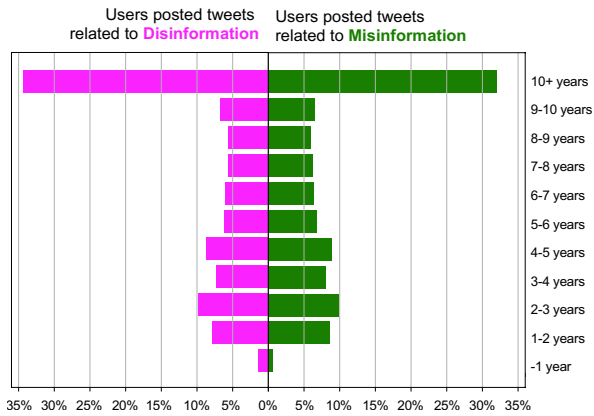}
    \caption{The distribution of the time that elapsed since the user account creation date from two perspectives: disinformation and misinformation}
    \label{example_year}
\end{figure}

\subsection{User profiles}
We aim to analyze the users who spread false information.
False news dissemination processes and user information are effective for fake news detection and understanding the formation of an echo chamber cycle~\cite{del2016echo}.

Figure~\ref{follower} shows the distribution of the count of followers and followees of 20,000 users, who were randomly selected from users who posted news stories labeled as disinformation or misinformation.
Users who spread disinformation tend to have more followers than those who spread misinformation.
The follower and followee counts of the users generally follow a power-law distribution, which is commonly observed in social network structures.
There is a spike around 5,000 in the followee count distribution for both, owing to Twitter restrictions.

Figure~\ref{example_year} shows the distribution of the time that has elapsed since each user created their account.
The distributions for disinformation and misinformation are similar.
However, when compared with reports on the distribution of users that spread false news in the US~\cite{shu2020fakenewsnet}, our results exhibit two features.
One is that few users have created accounts less than a year ago.
Another feature is that users who had been using Twitter for more than 10 years accounted for a large portion of disinformation and misinformation disseminators.
We believe that these characteristics are due to the fact that social media ``bot accounts'' are less active in Japan than in the US.

Finally, we investigated the ratio between ``bot accounts'' and human users that were involved in tweets related to misinformation and disinformation.
We randomly selected 10,000 users from each category and performed bot detection using the Botometer API~\cite{davis2016botornot}.
As a result, the ratio of ``bot accounts'' to human users is similar in the two categories: approximately 8\% for disinformation and 6\% for misinformation.
However, a comparison of reports on the ratio of bot users that spread false news in the US~\cite{shu2020fakenewsnet} and Japan shows that there are fewer bot users in Japan.
Specifically, almost 22\% of users that disseminate false news are bots in the US, whereas the corresponding percentage for Japanese users is less than 10\%.

\section{Conclusion and Future Work}
We proposed a novel annotation scheme to capture false news from various perspectives based on our investigations of the definition of ``fake news'' and existing fake news detection datasets.
We expect to reach an in-depth understanding of the phenomenon of ``fake news’’ using our annotation scheme, which utilizes fine-grained labeling that incorporates intent, negative social impact, targeting, and uniform labeling, and extends beyond factuality.
Subsequently, the first Japanese fake news dataset was constructed based on the annotation scheme to facilitate the study of fake news in Japan.

However, our Japanese fake news dataset is limited by a small sample size, owing to the small number of fact-checking articles that have been created.
To mitigate this limitation, we will continue to expand the dataset.
Furthermore, we will use our annotation scheme to construct datasets for false news in other languages.
These research endeavors will enable the following future studies:
\begin{itemize}
    \item An examination of the extent to which existing fake news detection models and language models can classify the labels assigned to our annotation scheme from social media posts.
    This is an important task to understand the limits of what machine learning methods can automatically classify.
    Although we were not able to conduct this investigation, owing to the small sample size of our constructed dataset in this study, we can proceed with this investigation in future work by extending our dataset.
    \item A comparison of the characteristics of fake news, such as the linguistic patterns and its diffusion patterns, across multiple countries.
    This task has rarely been performed.
    Section 5.3 compared statistics on the users who spread fake news from the existing English fake news dataset and our dataset.
    The construction of fake news datasets in other languages using the same scheme can enable multi-country comparisons on a more extensive scale.
    This may help us discover the unknown properties of fake news.
\end{itemize}


\section*{Bibliographical References}\label{reference}

\bibliographystyle{lrec2022-bib}
\bibliography{ref}


\end{document}